# From Conditional Oughts to Qualitative Decision Theory

Judea Pearl
Cognitive Systems Laboratory
University of California, Los Angeles, CA 90024
*judea@cs.ucla.edu*
*Content Areas: Commonsense Reasoning,
Probabilistic Reasoning, Reasoning about Action*

## Abstract

The primary theme of this investigation is a decision theoretic account of conditional ought statements (e.g., "You ought to do $A$, if $C$") that rectifies glaring deficiencies in classical deontic logic. The resulting account forms a sound basis for *qualitative decision theory*, thus providing a framework for qualitative planning under uncertainty. In particular, we show that adding causal relationships (in the form of a single graph) as part of an epistemic state is sufficient to facilitate the analysis of action sequences, their consequences, their interaction with observations, their expected utilities and, hence, the synthesis of plans and strategies under uncertainty.

## 1  INTRODUCTION

In natural discourse, "ought" statements reflect two kinds of considerations: requirements to act in accordance with moral convictions or peer's expectations, and requirements to act in the interest of one's survival, namely, to avoid danger and pursue safety. Statements of the second variety are natural candidates for decision theoretic analysis, albeit qualitative in nature, and these will be the focus of our discussion. The idea is simple. A sentence of the form "You ought to do $A$ if $C$" is interpreted as shorthand for a more elaborate sentence: "If you observe, believe, or know $C$, then the expected utility resulting from doing $A$ is much higher than that resulting from not doing $A$".[1] The longer sentence combines several modalities that have been the subjects of AI investigations: observation, belief, knowledge, probability ("expected"), desirability ("utility"), causation ("resulting from"), and, of course, action ("doing $A$"). With the exception of utility, these modalities have been formulated recently using qualitative, order-of-magnitude abstractions of probability theory (Goldszmidt & Pearl 1992, Goldszmidt 1992). Utility preferences themselves, we know from decision theory, can be fairly unstructured, save for obeying asymmetry and transitivity. Thus, paralleling the order-of-magnitude abstraction of probabilities, it is reasonable to score consequences on an integer scale of utility: very desirable ($U = O(1/\epsilon)$), very undesirable ($U = -O(1/\epsilon)$), bearable ($U = O(1)$), and so on, mapping each linguistic assessment into the appropriate $\pm O(1/\epsilon^i)$ utility rating. This utility rating, when combined with the infinitesimal rating of probabilistic beliefs (Goldszmidt & Pearl 1992), should permit us to rate actions by the expected utility of their consequences, and a requirement to do $A$ would then be asserted iff the rating of doing $A$ is substantially (i.e., a factor of $1/\epsilon$) higher than that of not doing $A$.

This decision theoretic agenda, although conceptually straightforward, encounters some subtle difficulties in practice. First, when we deal with actions and consequences, we must resort to causal knowledge of the domain and we must decide how such knowledge is to be encoded, organized, and utilized. Second, while theories of actions are normally formulated as theories of temporal changes (Shoham 1988, Dean & Kanazawa 1989), ought statements invariably suppress explicit references to time, strongly suggesting that temporal information is redundant, namely, it can be reconstructed if required, but glossed over otherwise. In other words, the fact that people comprehend, evaluate and follow non-temporal ought statements suggests that people adhere to some canonical, yet implicit assumptions about temporal progression of events, and that no account can be complete without making these assumptions explicit. Third, actions in decision theory predesignated explicitly to a few distinguished atomic variables, while statements of the type "You ought to do $A$" are presumed applicable to any arbitrary proposition $A$.[2] Finally, decision theoretic methods, especially those based on static influence diagrams, treat both the informational relationships between observations and actions and the causal relationships between actions and consequences as instantaneous (Chapter 6, Shachter 1986, Pearl 1988). In reality, the effect of our next action might be to invalidate currently observed properties, hence any non-temporal account of ought must carefully distinguish properties that are influenced by the action from those that will persist despite the action, and must explicate therefore some canonical assumptions about persistence.

---

[1] An alternative interpretation, in which "doing $A$" is required to be substantially superior to both "not doing $A$" and "doing not-$A$" is equally valid, and could be formulated as a straightforward extension of our analysis.

[2] This has been an overriding assumption in both the deontic logic and the preference logic literatures.

These issues are the primary focus of this paper. We start by presenting a brief introduction to infinitesimal probabilities and showing how actions, beliefs, and causal relationships are represented by ranking functions $\kappa(\omega)$ and causal networks $\Gamma$ (Section 2). In Section 3 we present a summary of the formal results obtained in this paper, including an assertability criterion for conditional oughts. Sections 4 and 5 explicate the assumptions leading to the criterion presented in Section 3. In Section 4 we introduce an integer-valued utility ranking $\mu(\omega)$ and show how the three components, $\kappa(\omega), \Gamma$, and $\mu(\omega)$, permit us to calculate, semi-qualitatively, the utility of an arbitrary proposition $\varphi$, the utility of a given action $A$, and whether we ought to do $A$. Section 5 introduces conditional oughts, namely, statements in which the action is contingent upon observing a condition $C$. A calculus is then developed for transforming the conditional ranking $\kappa(\omega|C)$ into a new ranking $\kappa_A(\omega|C)$, representing the beliefs an agent will possess after implementing action $A$, having observed $C$. These two ranking functions are then combined with $\mu(\omega)$ to form an assertability criterion for the conditional statement $O(A|C)$: "We ought to do $A$, given $C$". In Section 6 we compare our formulation to other accounts of ought statements, in particular deontic logic, preference logic, counterfactual conditionals, and quantitative decision theory.

## 2 INFINITESIMAL PROBABILITIES, RANKING FUNCTIONS, CAUSAL NETWORKS, AND ACTIONS

1. (*Ranking Functions*). Let $\Omega$ be a set of worlds, each world $\omega \in \Omega$ being a truth-value assignment to a finite set of atomic variables $(X_1, X_2, ..., X_n)$ which in this paper we assume to be bi-valued, namely, $X_i \in \{true, false\}$. A belief *ranking function* $\kappa(\omega)$ is an assignment of non-negative integers to the elements of $\Omega$ such that $\kappa(\omega) = 0$ for at least one $\omega \in \Omega$. Intuitively, $\kappa(\omega)$ represents the degree of surprise associated with finding a world $\omega$ realized, and worlds assigned $\kappa = 0$ are considered serious possibilities. $\kappa(\omega)$ can be considered an order-of-magnitude approximation of a probability function $P(\omega)$ by writing $P(\omega)$ as a polynomial of some small quantity $\epsilon$ and taking the most significant term of that polynomial, i.e.,

$$P(\omega) \cong C\epsilon^{\kappa(\omega)} \qquad (1)$$

Treating $\epsilon$ as an infinitesimal quantity induces a conditional ranking function $\kappa(\varphi|\psi)$ on propositions which is governed by Spohn's calculus (Spohn 1988):

$$\kappa(\Omega) = 0$$

$$\kappa(\varphi) = \begin{cases} \min_\omega \kappa(\omega) & \text{for } \omega \models \varphi \\ \infty & \text{for } \omega \models \neg\varphi \end{cases}$$

$$\kappa(\varphi|\psi) = \kappa(\varphi \wedge \psi) - \kappa(\psi) \qquad (2)$$

2. (*Stratified Rankings and Causal Networks* (Goldszmidt & Pearl 1992)). A *causal network* is a directed acyclic graph (dag) in which each node corresponds to an atomic variable and each edge $X_i \longrightarrow X_j$ asserts that $X_i$ has a direct causal influence on $X_j$. Such networks provide a convenient data structure for encoding two types of information: how the initial ranking function $\kappa(\omega)$ is formed, and how external actions would influence the agent's belief ranking $\kappa(\omega)$. Formally, causal networks are defined in terms of two notions: stratification and actions.

A ranking function $\kappa(\omega)$ is said to be *stratified* relative to a dag $\Gamma$ if

$$\kappa(\omega) = \sum_i \kappa(X_i(\omega)|\mathbf{pa}_i(\omega)) \qquad (3)$$

where $\mathbf{pa}_i(\omega)$ are the parents of $X_i$ in $\Gamma$ evaluated at state $\omega$. Given a ranking function $\kappa(\omega)$, any edge-minimal dag $\Gamma$ satisfying Eq. (3), is called a *Bayesian network* of $\kappa(\omega)$ (Pearl 1988). A dag $\Gamma$ is said to be a causal network of $\kappa(\omega)$ if it is a Bayesian network of $\kappa(\omega)$ and, in addition, it admits the following representation of actions.

3. (*Actions*) The effect of an atomic action $do(X_i = true)$ is represented by adding to $\Gamma$ a link $DO_i \longrightarrow X_i$, where $DO_i$ is a new variable taking values in $\{do(x_i), do(\neg x_i), idle\}$ and $x_i$ stands for $X_i = true$. Thus, the new parent set of $X_i$ is $\mathbf{pa}'_i = \mathbf{pa}_i \cup \{DO_i\}$ and it is related to $X_i$ by

$$\kappa(X_i(\omega)|\mathbf{pa}'_i(\omega)) = \begin{cases} \kappa(X_i(\omega)|\mathbf{pa}_i(\omega)) & \text{if } DO_i = idle \\ \infty & \text{if } DO_i = do(y) \text{ and } X_i(\omega) \neq y \\ 0 & \text{if } DO_i = do(y) \text{ and } X_i(\omega) = y \end{cases} \qquad (4)$$

The effect of performing action $do(x_i)$ is to transform $\kappa(\omega)$ into a new belief ranking, $\kappa_{x_i}(\omega)$, given by

$$\kappa_{x_i}(\omega) = \begin{cases} \kappa'(\omega|do(x_i)) & \text{for } \omega \models x_i \\ \infty & \text{for } \omega \models \neg x_i \end{cases} \qquad (5)$$

where $\kappa'$ is the ranking dictated by the augmented network $\Gamma \cup \{DO_i \longrightarrow X_i\}$ and Eqs. (3) and (4).

This representation embodies the following aspects of actions:

(i) An action $do(x_i)$ can affect only the descendants of $X_i$ in $\Gamma$.

(ii) Fixing the value of $\mathbf{pa}_i$ (by some appropriate choice of actions) renders $X_i$ unaffected by any external intervention $do(x_\kappa), \kappa \neq i$.

## 3   SUMMARY OF RESULTS

The assertability condition we are about to develop in this paper requires the specification of an epistemic state $ES = (\kappa(\omega), \Gamma, \mu(\omega))$ which consists of three components:

$\kappa(\omega)$ - an ordinal belief ranking function on $\Omega$.

$\Gamma$ - a causal network of $\kappa(\omega)$.

$\mu(\omega)$ - an integer-valued utility ranking of worlds, where $\mu(\omega) = \pm i$ assigns to $\omega$ a utility $U(\omega) = \pm O(1/\epsilon^i), i = 0, 1, 2, ...$.

The main results of this paper can be summarized as follows:

1. Let $W_i^+$ and $W_i^-$ be the formulas whose models receive utility ranking $+i$ and $-i$, respectively, and let $\kappa'(\omega)$ denote the ranking function that prevails after establishing the truth of event $\varphi$, where $\varphi$ is an arbitrary proposition (i.e., $\kappa'(\neg\varphi) = \infty$ and $\kappa'(\varphi) = 0$). The expected utility rank of $\varphi$ is characterized by two integers

$$\begin{aligned} n^+ &= \max_i[0; \ i - \kappa'(W_i^+ \wedge \varphi)] \\ n^- &= \max_i[0; \ i - \kappa'(W_i^- \wedge \varphi)] \end{aligned} \quad (6)$$

and is given by

$$\mu[(\varphi; \kappa'(\omega)] = \begin{cases} \text{ambiguous} & \text{if } n^+ = n^- > 0 \\ n^+ - n^- & \text{otherwise} \end{cases} \quad (7)$$

2. A conditional ought statement $O(A|C)$ is assertable in $ES$ iff

$$\mu(A; \kappa_A(\omega|C)) > \mu(true; \kappa(\omega|C)) \quad (8)$$

where $A$ and $C$ are arbitrary propositions and the ranking $\kappa_A(\omega|C)$ (to be defined in step 3) represents the beliefs that an agent anticipates holding, after implementing action $A$, having observed $C$.

3. If $A$ is a conjunction of atomic propositions, $A = \bigwedge_{j \in J} A_j$, where each $A_j$ stands for either $X_j = true$ or $X_j = false$, then the post-action ranking $\kappa_A(\omega|C)$ is given by the formula

$$\kappa_A(\omega|C) = \kappa(\omega) - \sum_{i \in J \cup R} \kappa(X_i(\omega)|\mathbf{pa}_i(\omega)) + \min_{\omega'}[\sum_{i \notin J} S_i(\omega, \omega') + \kappa(\omega'|C)] \quad (9)$$

where $R$ is the set of root nodes and

$$S_i(\omega, \omega') = \begin{cases} s_i & \text{if } X_i(\omega) \neq X_i(\omega') \text{ and } \mathbf{pa}_i = \emptyset \\ s_i & \text{if } X_i(\omega) \neq X_i(\omega'), \mathbf{pa}_i \neq \emptyset \text{ and} \\ & \kappa(\neg X_i(\omega)|\mathbf{pa}_i(\omega)) = 0 \\ 0 & \text{otherwise} \end{cases} \quad (10)$$

$S(\omega, \omega')$ represents persistence assumptions: It is surprising (to degree $s_i \geq 1$) to find $X_i$ change from its pre-action value of $X_i(\omega')$ to a post-action value of $X_i(\omega)$ if there is no causal reason for the change.

If $A$ is a disjunction of actions, $A = \bigvee_l A^l$, where each $A^l$ is a conjunction of atomic propositions, then

$$\kappa_A(\omega|C) = \min_l \kappa_{A^l}(\omega|C) \quad (11)$$

## 4   FROM UTILITIES AND BELIEFS TO GOALS AND ACTIONS

Given a proposition $\varphi$ that describes some condition or event in the world, what information is needed before we can evaluate the merit of obtaining $\varphi$, or, at the least, whether $\varphi_1$ is "preferred" to $\varphi_2$? Clearly, if we are to apply the expected utility criterion, we should define two measures on possible worlds, a probability measure $P(\omega)$ and a utility measure $U(\omega)$. The first rates the likelihood that a world $\omega$ will be realized, while the second measures the desirability of $\omega$. Unfortunately, probabilities and utilities in themselves are not sufficient for determining preferences among propositions. The merit of obtaining $\varphi$ depends on at least two other factors: how the truth of $\varphi$ is established, and what control we possess over which model of $\varphi$ will eventually prevail. We will demonstrate these two factors by example.

Consider the proposition $\varphi =$ "The ground is wet". In the midst of a drought, the consequences of this statement would depend critically on whether we watered the ground (action) or we happened to find the ground wet (observation). Thus, the utility of a proposition $\varphi$ clearly depends on how we came to know $\varphi$, by mere observation or by willful action. In the first case, finding $\varphi$ true may provide information about the natural process that led to the observation $\varphi$, and we should change the current probability from $P(\omega)$ to $P(\omega|\varphi)$. In the second case, our actions may perturb the natural flow of events, and $P(\omega)$ will change without shedding light on the typical causes of $\varphi$. We will denote the probability resulting from externally enforcing the truth of $\varphi$ by $P_\varphi(\omega)$, which will be further explicated in Section 5 in terms of the causal network $\Gamma$.[3]

However, regardless of whether the probability function $P(\omega|\varphi)$ or $P_\varphi(\omega)$ results from learning $\varphi$, we are still unable to evaluate the merit of $\varphi$ unless we understand what control we have over the opportunities offered by $\varphi$. Simply taking the expected utility $U(\varphi) = \Sigma_\omega [P(\omega|\varphi) U(\omega)]$ amounts to assuming that the agent is to remain totally passive until Nature selects a world $\omega$ with probability $P(\omega|\varphi)$, as in a game of chance. It ignores subsequent actions which the agent might be able to take so as to change this probability. For example, event $\varphi$ might provide the agent with the option of conducting further tests so as to determine with greater certainty which world would eventually be realized. Likewise, in case $\varphi$ stands for "Joe went to get his gun", our agent might possess the wisdom to protect itself by escaping in the next taxicab.

---

[3]The difference between $P(\omega|\varphi)$ and $P_\varphi(\omega)$ is precisely the difference between conditioning and "imaging" (Lewis 1973), and between belief revision and belief update (Alchourron et.al. 1985, Katsuno & Mendelzon 1991, Goldszmidt & Pearl 1992). It also accounts for the difference between indicative and subjunctive conditionals – a topic of much philosophical discussion (Harper et.al. 1980).

In practical decision analysis the utility of being in a situation $\varphi$ is computed using a dynamic programming approach, which assumes that subsequent to realizing $\varphi$ the agent will select the optimal sequence of actions from those enabled by $\varphi$. This computation is rather exhaustive and is often governed by some form of myopic approximation (Chapter 6, Pearl 1988). Ought statements normally refer to a single action $A$, tacitly assuming that the choice of subsequent actions, if available, is rather obvious and their consequences are well understood. We say, for example, "You ought to get some food", assuming that the food would subsequently be eaten and not be left to rot in the car. In our analysis, we will make a similar myopic approximation, assuming either that action $A$ is terminal or that the consequences of subsequent actions (if available) are already embodied in the functions $P(\omega)$ and $\mu(\omega)$. We should keep in mind, however, that the result of this myopic approximation is not applicable to *all* actions; in sequential planning situations, some actions may be selected for the sole purpose of enabling certain subsequent actions.

Denote by $P'(\omega)$ the probability function that would prevail after obtaining $\varphi$.[4] Let us examine next how the expected utility criterion $U(\varphi) = \Sigma P'(\omega) U(\omega)$ translates into the language of ranking functions.

Let us assume that $U$ takes on values in $\{-O(1/\epsilon), O(1), +O(1/\epsilon)\}$, read as {very undesirable, bearable, very desirable}. For notational simplicity, we can describe these linguistic labels as a *utility ranking* function $\mu(\omega)$ that takes on the values $-1$, $0$, and $+1$, respectively. Our task, then, is to evaluate the rank $\mu(\varphi)$, as dictated by the expected value of $U(\omega)$ over the models of $\varphi$.

Let the sets of worlds assigned the ranks $-1$, $0$, and $+1$ be represented by the formulas $W^-$, $W^0$, and $W^+$, respectively, and let the intersections of these sets with $\varphi$ be represented by the formulas $\varphi^-$, $\varphi^0$, and $\varphi^+$, respectively. The expected utility of $\varphi$ is given by $-C_-/\epsilon\ P'(W^-) + C_0\ P'(W^0) + C_+/\epsilon\ P'(W^+)$, where $C_-, C_0$, and $C_+$ are some positive coefficients. Introducing now the infinitesimal approximation for $P'$, in the form of the ranking function $\kappa'$, we obtain

$$U(\varphi) = \begin{cases} -O(1/\epsilon) & \text{if } \kappa'(\varphi^-) = 0 \\ & \quad \text{and } \kappa'(\varphi^+) > 0 \\ O(1) & \text{if } \kappa'(\varphi^-) > 0 \\ & \quad \text{and } \kappa'(\varphi^+) > 0 \\ +O(1/\epsilon) & \text{if } \kappa'(\varphi^-) > 0 \\ & \quad \text{and } \kappa'(\varphi^+) = 0 \\ \text{ambiguous} & \text{if } \kappa'(\varphi^-) = 0 \end{cases} \quad (12)$$

The ambiguous status reflects a state of conflict $U(\varphi) = -C_-/\epsilon + C_+/\epsilon$, where there is a serious possibility of ending in either terrible disaster or enormous success. Recognizing that ought statements are often intended to avert such situations (e.g., "You ought to invest in something safer"), we may take a risk-averse attitude and rank ambiguous states as low as $U = -O(1/\epsilon)$ (other attitudes are, of course, perfectly legitimate). This attitude, together with $\kappa'(\varphi) = 0$, yields the desired expression for $\mu(\varphi;\ \kappa'(\omega))$:

$$\mu(\varphi; \kappa'(\omega)) = \begin{cases} -1 & \text{if } \kappa'(W^-|\varphi) = 0 \\ 0 & \text{if } \kappa'(W^- \vee W^+|\varphi) > 0 \\ +1 & \text{if } \kappa'(W^-|\varphi) > 0 \\ & \quad \text{and } \kappa'(W^+|\varphi) = 0 \end{cases} \quad (13)$$

---
[4] $P'(\omega) = P(\omega|\varphi)$ in case $\varphi$ is observed, and $P'(\omega) = P_\varphi(\omega)$ in case $\varphi$ is enacted. In both cases $P'(\varphi) = 1$.

The three-level utility model is, of course, only a coarse rating of desirability. In a multi-level model, where $W_i^+$ and $W_i^-$ are the formulas whose models receive utility ranking $+i$ and $-i$, respectively[5], and $i = 0, 1, 2, ...$, the ranking of the expected utility of $\varphi$ is given by Eq. (7) (Section 3).

Having derived a formula for the utility rank of an arbitrary proposition $\varphi$, we are now in a position to formulate our interpretation of the deontic expression $O(A|C)$: "You ought to do $A$ if $C$, iff the expected utility associated with doing $A$ is much higher than that associated with not doing $A$". We start with a belief ranking $\kappa(\omega)$ and a utility ranking $\mu(\omega)$, and we wish to assess the utilities associated with doing $A$ versus not doing $A$, given that we observe $C$. The observation $C$ would transform our current $\kappa(\omega)$ into $\kappa(\omega|C)$. Doing $A$ would further transform $\kappa(\omega|C)$ into $\kappa'(\omega) = \kappa_A(\omega|C)$, while not doing $A$ would render $\kappa(\omega|C)$ unaltered, so $\kappa'(\omega) = \kappa(\omega|C)$. Thus, the utility rank associated with doing $A$ is given by $\mu(A; \kappa'_A(\omega|C))$, while that associated with not doing $A$ is given by $\mu(C; \kappa(\omega|C)) = \mu(true; \kappa(\omega|C))$. Consequently, we can write the assertability criterion for conditional ought as

$$O(A|C) \quad \text{iff} \quad \mu(A; \kappa_A(\omega|C)) > \mu(true; \kappa(\omega|C)) \quad (14)$$

where the function $\mu(\varphi; \kappa(\omega))$ is given in Eq. (13).

We remark that the transformation from $\kappa(\omega|C)$ to $\kappa_A(\omega|C)$ requires causal knowledge of the domain, which will be provided by the causal network $\Gamma$ (Section 5). Once we are given $\Gamma$ it will be convenient to encode both $\kappa(\omega)$ and $\mu(\omega)$ using integer-valued labels on the links of $\Gamma$. Moreover, it is straightforward to apply Eqs. (7) and (14) to the usual decision theoretic tasks of selecting an optimal action or an optimal information-gathering strategy (Chapter 6, Pearl 1988).

**Example 1:**
To demonstrate the use of Eq. (14), let us examine the assertability of "If it is cloudy you ought to take an umbrella" (Boutilier 1993). We assume three atomic propositions, $c$ - "Cloudy", $r$ - "Rain", and $u$ - "Having an Umbrella", which form eight worlds, each corresponding to a complete truth assignment to $c$, $r$, and $u$.

---
[5] In practice, the specification of $U(\omega)$ is done by defining an integer-valued variable $V$ (connoting "value") as a function of a select set of atomic variables. $W^i$ would correspond then to the assertion $V = i$, $i = 0, 1, 2, ...$.

To express our belief that rain does not normally occur in a clear day, we assign a $\kappa$ value of 1 (indicating one unit of surprise) to any world satisfying $r \wedge \neg c$ and a $\kappa$ value of 0 to all other worlds (indicating a serious possibility that any such world may be realized). To express the fear of finding ourselves in the rain without an umbrella, we assign a $\mu$ value of $-1$ to worlds satisfying $r \wedge \neg u$ and a $\mu$ value of 0 to all other worlds. Thus, $W^+ = false$, $W^0 = \neg(r \wedge \neg u)$, and $W^- = r \wedge \neg u$.

In this simple example, there is no difference between $\kappa_A(\omega)$ and $\kappa(\omega|A)$ because the act $A =$ "Taking an umbrella" has the same implications as the finding "Having an umbrella". Thus, to evaluate the two expressions in Eq. (14), with $A = u$ and $C = c$, we first note that

$$\kappa(W^-|u,c) = \kappa(r \wedge \neg u|u,c) = \infty$$
$$\kappa(W^- \vee W^+|u,c) = \infty$$

so
$$\mu(u; \kappa(\omega|u,c)) = 0$$

Similarly,
$$\kappa(W^-|c) = \kappa(r \wedge \neg u|c) = 0$$

hence
$$\mu(c; \kappa(\omega|c)) = -1 \qquad (15)$$

Thus, $O(u|c)$ is assertable according to the criterion of Eq. (14).

Note that although $\kappa(\omega)$ does not assume that normally we do not have an umbrella with us ($\kappa(u) > 0$), the advice to take an umbrella is still assertable, since leaving $u$ to pure chance might result in harsh consequences (if it rains).

Using the same procedure, it is easy to show that the example also sanctions the assertability of $O(\neg r|c, \neg u)$, which stands for "If it is cloudy and you don't have an umbrella, then you ought to undo (or stop) the rain". This is certainly useless advice, as it does not take into account one's inability to control the weather. Controllability information is not encoded in the ranking functions $\kappa$ and $\mu$; it should be part of one's causal theory and can be encoded in the language of causal networks using costly preconditions that, until satisfied, would forbid the action $do(A)$ from having any effect on $A$.[6]

## 5  COMBINING ACTIONS AND OBSERVATIONS

In this section we develop a probabilistic account for the term $\kappa_A(\omega|C)$, which stands for the belief ranking that would prevail if we act $A$ after observing $C$, i.e., the $A$-update of $k(\omega|C)$. First we note that this update cannot be obtained by simply applying the update formula developed in (Eq. (2.2), Goldszmidt & Pearl 1992),

$$\kappa_A(\omega) = \begin{cases} \kappa(\omega) - \kappa(A|\mathbf{pa}_A(\omega)) & \omega \models A \\ \infty & \omega \models \neg A \end{cases} \qquad (16)$$

where $\mathbf{pa}_A(\omega)$ are the parents (or immediate causes) of $A$ in the causal network $\Gamma$ evaluated at $\omega$. The formula above was derived under the assumption that $\Gamma$ is not loaded with any observations (e.g., $C$) and renders $\kappa_A(\omega)$ undefined for worlds $\omega$ that are excluded by previous observations and reinstated by $A$.

---

[6]In decision theory it is customary to attribute direct costs to actions, which renders $\mu(\omega)$ action-dependent. An alternative, which is more convenient when actions are not enumerated explicitly, is to attribute costs to preconditions that must be satisfied before (any) action becomes effective.

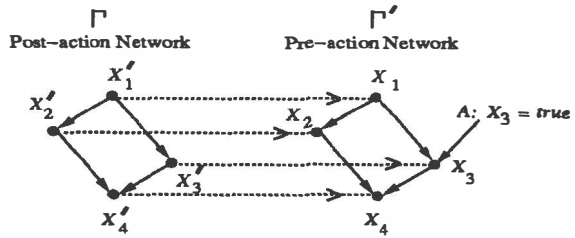

Figure 1: Persistence interactions between two causal networks

To motivate the proper transformation from $\kappa(\omega)$ to $\kappa_A(\omega|C)$, we consider two causal networks, $\Gamma'$ and $\Gamma$ respectively representing the agent's epistemic states before and after the action (see Figure 1). Although the structures of the two networks are almost the same ($\Gamma$ contains additional root nodes representing the action $do(A)$), it is the interactions between the corresponding variables that determine which beliefs are going to persist in $\Gamma$ and which are to be "clipped" by the influence of action $A$.

Let every variable $X_i'$ in $\Gamma'$ be connected to the corresponding variable $X_i$ in $\Gamma$ by a directed link $X_i' \longrightarrow X_i$ that represents persistence by default, namely, the natural tendency of properties to persist, unless there is a cause for a change. Thus, the parent set of each $X_i$ in $\Gamma$ has been augmented with one more variable: $X_i'$. To specify the conditional probability of $X_i$, given its new parent set $\{\mathbf{pa}_{X_i} \cup X_i'\}$, we need to balance the tendency of $X_i$ to persist (i.e., be equal to $X_i'$) against its tendency to obey the causal influence exerted by $\mathbf{pa}_{X_i}$. We will assume that persistence forces yield to causal forces and will perpetuate only those properties that are not under any causal influence to terminate. In terms of ranking functions, this assumption reads:

$\kappa(X_i(\omega)|\mathbf{pa}_i(\omega), X_i'(\omega')) =$
$$\begin{cases} s_i & \text{if } \mathbf{pa}_i = \emptyset \text{ and } X_i(\omega) \neq X_i(\omega') \\ s_i + \kappa(X_i(\omega)|\mathbf{pa}_i(\omega)) & \text{if } X_i(\omega) \neq X_i'(\omega') \text{ and} \\ & \qquad \kappa(\neg X_i(\omega)|\mathbf{pa}_i(\omega)) = 0 \\ \kappa(X_i(\omega)|\mathbf{pa}_i(\omega)) & \text{otherwise} \end{cases} \qquad (17)$$

where $\omega'$ and $\omega$ specify the truth values of the variables in the corresponding networks, $\Gamma'$ and $\Gamma$, and $s_i \geq 1$ is a constant characterizing the tendency of $X_i$ to persist. Eq. (17) states that the past value of $X_i$ may affect the normal relation between $X_i$ and its parents only when it differs from the current value and, at the same time, the parents of $X_i$ do not compel the change. In such a case, the inequality $X_i(\omega) \neq X'_i(\omega')$ contributes $s_i$ units of surprise to the normal relation between $X_i$ and its parents.[7] The unique feature of this model, unlike the one proposed in (Goldszmidt & Pearl 1992), is that persistence defaults can be violated by causal factors without forcing us to conclude that such factors are abnormal.

Eq. (17) specifies the conditional rank $\kappa(X|\mathbf{pa}_X)$ for every variable $X$ in the combined networks and, hence, it provides a complete specification of the joint rank $\kappa(\omega, \omega')$.[8] The desired expression for the post-action ranking $\kappa_A(\omega)$ can then be obtained by marginalizing $\kappa(\omega, \omega')$ over $\omega'$:

$$\kappa_A(\omega) = \min_{\omega'} \kappa(\omega, \omega') \tag{18}$$

We need, however, to account for the fact that some variables in network $\Gamma$ are under the direct influence of the action $A$, and hence the parents of these nodes are replaced by the action node $do(A)$. If $A$ consists of a conjunction of atomic propositions, $A = \wedge_{j \in J} A_j$, where each $A_j$ stands for either $X_j = true$ or $X_j = false$, then each $X_i$, $i \in J$, should be exempt from incurring the spontaneity penalty specified in Eq. (17). Additionally, in calculating $\kappa(\omega, \omega')$ we need to sum $\kappa(X_i(\omega)|\mathbf{pa}_i(\omega), X'_i(\omega'))$ only over $i \notin J$, namely, over variables not under the direct influence of $A$. Thus, collecting terms and writing $\kappa(\omega) = \sum_i \kappa(X_i(\omega)|\mathbf{pa}_i(\omega))$, we obtain

$$\kappa_A(\omega|C) = \kappa(\omega) - \sum_{i \in J \cup R} \kappa(X_i(\omega)|\mathbf{pa}_i(\omega)) + \min_{\omega'}[\sum_{i \notin J} S_i(\omega, \omega') + \kappa(\omega'|C)] \tag{19}$$

where $R$ is the set of root nodes and

$$S_i(\omega, \omega') = \begin{cases} s_i & \text{if } X_i(\omega) \neq X_i(\omega') \text{ and } \mathbf{pa}_i = \emptyset \\ s_i & \text{if } X_i(\omega) \neq X_i(\omega'), \mathbf{pa}_i \neq \emptyset \text{ and } \\ & \kappa(\neg X_i(\omega)|\mathbf{pa}_i(\omega)) = 0 \\ 0 & \text{otherwise} \end{cases} \tag{20}$$

---

[7] This is essentially the persistence model used by Dean and Kanazawa (Dean & Kanazawa 1989), in which $s_i$ represents the survival function of $X_i$. The use of ranking functions allows us to distinguish crisply between changes that are causally supported, $\kappa(\neg X_i(\omega)|\mathbf{pa}_i(\omega)) > 0$, and those that are unsupported, $\kappa(\neg X_i(\omega)|\mathbf{pa}_i(\omega)) = 0$.

[8] The expressions, familiar in probability theory,

$$P(\omega, \omega') = \prod_j P(X_j(\omega, \omega')|\mathbf{pa}_j(\omega, \omega')), \quad P(\omega) = \sum_{\omega'} P(\omega, \omega')$$

translate into the ranking expressions

$$\kappa(\omega, \omega') = \sum_j \kappa(X_j(\omega, \omega'))|\mathbf{pa}_j(\omega, \omega')), \quad \kappa(\omega) = \min_{\omega'} \kappa(\omega, \omega')$$

where $j$ ranges over all variables in the two networks.

Eq. (19) demonstrates that the effect of observations and actions can be computed as an updating operation on epistemic states, these states being organized by a fixed causal network, with the only varying element being $\kappa$, the belief ranking. Long streams of observations and actions could therefore be processed as a sequence of updates on some initial state, without requiring analysis of long chains of temporally indexed networks, as in Dean and Kanazawa (1989).

To handle disjunctive actions such as "Paint the wall either red or blue" one must decide between two interpretations: "Paint the wall red or blue regardless of its current color" or "Paint the wall either red or blue but, if possible, do not change its current color" (see Katsuno & Mendelzon 1991 and Goldszmidt & Pearl 1992). We will adopt the former interpretation, according to which "$do(A \vee B)$" is merely a shorthand for "$do(A) \vee do(B)$". This interpretation is particularly convenient for ranking systems, because for any two propositions, $A$ and $B$, we have

$$\kappa(A \vee B) = \min[\kappa(A); \kappa(B)] \tag{21}$$

Thus, if we do not know which action, $A$ or $B$, will be implemented but consider either to be a serious possibility, then

$$\kappa_{A \vee B}(\omega) = \min[\kappa_A(\omega); \kappa_B(\omega)] \tag{22}$$

Accordingly, if $A$ is a disjunction of actions, $A = \bigvee_l A^l$, where each $A^l$ is a conjunction of atomic propositions, then

$$\kappa_A(\omega|C) = \min_l \kappa_{A^l}(\omega|C) \tag{23}$$

**Example 2**

To demonstrate the interplay between actions and observations, we will test the assertability of the following dialogue:
Robot 1: It is too dark in here.
Robot 2: Then you ought to push the switch up.
Robot 1: The switch is already up.
Robot 2: Then you ought to push the switch down.

The challenge would be to explain the reversal of the "ought" statement in response to the new observation "The switch is already up". The inferences involved in this example revolve around identifying the type of switch Robot 1 is facing, that is whether it is normal ($n$) or abnormal ($\neg n$) (a normal switch is one that should be pushed up ($u$) to turn the light on ($l$)). The causal network, shown in Figure 2, involves three variables:

$L$ - the current state of the light ($l$ vs $\neg l$),
$S$ - the current position of the switch ($u$ vs $\neg u$), and
$T$ - the type of switch at hand ($n$ vs $\neg n$).

The variable $L$ stands in functional relationship to $S$ and $T$, via

$$l = (n \wedge u) \vee (\neg n \wedge \neg u) \tag{24}$$

or, equivalently, $k = \infty$ unless $l$ satisfies the relation above.

Since initially the switch is believed to be normal, we set $\kappa(\neg n) = 1$, resulting in the following initial ranking:

| $S$ | $T$ | $L$ | $\kappa(\omega)$ |
|---|---|---|---|
| $u$ | $n$ | $l$ | 0 |
| $\neg u$ | $n$ | $\neg l$ | 0 |
| $u$ | $\neg n$ | $\neg l$ | 1 |
| $\neg u$ | $\neg n$ | $l$ | 1 |

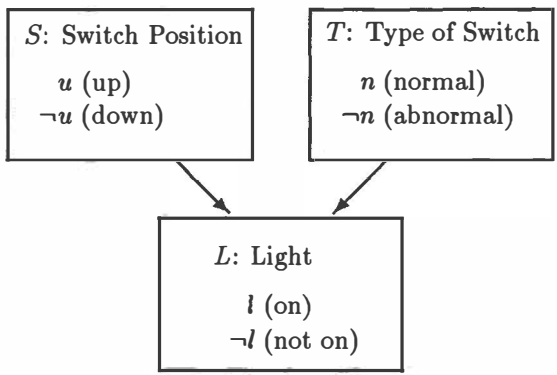

Figure 2: Causal network for Example 2

We also assume that Robot 1 prefers light to darkness, by setting

$$\mu(\omega) = \begin{cases} -1 & \text{if } \omega \models \neg l \\ 0 & \text{if } \omega \models l \end{cases} \quad (25)$$

The first statement of Robot 1 expresses an observation $C = \neg l$, yielding

$$\kappa(\omega|C) = \begin{cases} 0 & \text{for } \omega = \neg u \wedge n \wedge \neg l \\ 1 & \text{for } \omega = u \wedge \neg n \wedge \neg l \\ \infty & \text{for all other worlds} \end{cases} \quad (26)$$

To evaluate $\kappa_A(\omega|C)$ for $A = u$, we now invoke Eq. (19), using the spontaneity functions

$$\begin{aligned} S_T(\omega, \omega') &= 1 \text{ if } T(\omega) \neq T(\omega') \\ S_L(\omega, \omega') &= 0 \text{ if } L(\omega) \neq L(\omega') \end{aligned} \quad (27)$$

because $L(\omega)$, being functionally determined by $\mathbf{pa}_L(\omega)$ is exempt from conforming to persistence defaults. Moreover, for action $A = u$ we also have $\kappa(u|\mathbf{pa}_A) = \kappa(u) = 0$, hence

$$\kappa_A(\omega|C) = \kappa(\omega) - \kappa(T(\omega))$$
$$\min_{\omega' = \omega'_1, \omega'_2} \{I[T(\omega) \neq T(\omega')] + \kappa(\omega'|C)\},$$
$$\text{for } \omega = \omega_1, \omega_2 \quad (28)$$

where $I[p]$ equals 1 (or 0) if $p$ is *true* (or *false*), and

$$\begin{aligned} \omega_1 &= u \wedge n \wedge l & \omega'_1 &= \neg u \wedge n \wedge \neg l \\ \omega_2 &= u \wedge \neg n \wedge \neg l & \omega'_2 &= u \wedge \neg n \wedge \neg l \end{aligned} \quad (29)$$

All other worlds are excluded by either $A = u$ or $C = \neg l$.

Minimizing Eq. (19) over the two possible $\omega'$ worlds, yields

$$\kappa_A(\omega|C) = \begin{cases} 0 & \text{for } \omega = \omega_1 \\ 1 & \text{for } \omega = \omega_2 \end{cases} \quad (30)$$

We see that $\omega_2 = u \wedge \neg n \wedge \neg l$ is penalized with one unit of surprise for exhibiting an unexplained change in switch type (initially believed to be normal).

It is worth noting how $\omega_1$, which originally was ruled out (with $\kappa = \infty$) by the observation $\neg l$, is suddenly reinstated after taking the action $A = u$. In fact, Eq. (19) first restores all worlds to their original $\kappa(\omega)$ value and then adjusts their value in three steps. First it excludes worlds satisfying $\neg A$, then adjusts the $\kappa(\omega)$ of the remaining worlds by an amount $\kappa(A|\mathbf{pa}_A(\omega))$, and finally makes an additional adjustment for violation of persistence.

From Eqs. (26) and (28), we see that $\kappa_A(l|C) = 0 < \kappa(l|C) = \infty$, hence the action $A = u$ meets the assertability criterion of Eq. (14) and the first statement, "You ought to push the switch up", is justified. At this point, Robot 2 receives a new piece of evidence: $S = u$. As a result, $\kappa(\omega|\neg l)$ changes to $\kappa(\omega|\neg l, u)$ and the calculation of $\kappa_A(\omega|C)$ needs to be repeated with a new set of observations, $C = \neg l \wedge u$. Since $\kappa(\omega'|\neg l, u)$ permits only one possible world $\omega' = u \wedge \neg n \wedge \neg l$, the minimization of Eq. (19) can be skipped, yielding (for $A = \neg u$)

$$\kappa_A(\omega|C) = \begin{cases} 0 \text{ for } w = \neg u \wedge \neg n \wedge l \\ 1 \text{ for } w = \neg u \wedge n \wedge \neg l \end{cases} \quad (31)$$

which, in turn, justifies the opposite "ought" statement ("Then you ought to push the switch down"). Note that although finding a normal switch is less surprising than finding an abnormal switch, a spontaneous transition to such a state would violate persistence and is therefore penalized by obtaining a $\kappa$ of 1.

## 6 RELATIONS TO OTHER ACCOUNTS

### 6.1 DEONTIC AND PREFERENCE LOGICS

Ought statements of the pragmatic variety have been investigated in two branches of philosophy, deontic logic and preference logic. Surprisingly, despite an intense effort to establish a satisfactory account of "ought" statements (Von Wright 1963, Van Fraassen 1973, Lewis 1973), the literature of both logics is loaded with paradoxes and voids of principle. This raises the question of whether "ought" statements are destined to forever elude formalization or that the approach taken by deontic logicians has been misdirected. I believe the answer involves a combination of both.

Philosophers hoped to develop deontic logic as a separate branch of conditional logic, not as a synthetic amalgam of logics of belief, action, and causation. In other words, they have attempted to capture the meaning of "ought" using a single modal operator $O(\cdot)$, instead of exploring the couplings between "ought" and other modalities, such as belief, action, causation, and desire. The present paper shows that such an isolationistic strategy has little chance of succeeding. Whereas one can perhaps get by without explicit reference to desire, it is absolutely necessary to have both probabilistic knowledge about the effect of observations on the likelihood of events and causal knowledge about actions and their consequences.

We have seen in Section 3 that to ratify the sentence "Given $C$, you ought to do $A$", we need to know not merely the relative desirability of the worlds delineated by the propositions $A \wedge C$ and $\neg A \wedge C$, but also the feasibility or likelihood of reaching any one of those worlds in the future, after making our choice of $A$. We also saw that this likelihood depends critically on how $C$ is confirmed, by observation or by action. Since this information cannot be obtained from the logical content of $A$ and $C$, it is not surprising that "almost every principle which has been proposed as fundamental to a preference logic has been rejected by some other source" (Mullen 1979).

In fact, the decision theoretic account embodied in Eq. (14) can be used to generate counterexamples to most of the principles suggested in the literature, simply by selecting a combination of $\kappa$, $\mu$, and $\Gamma$ that defies the proposed principle. Since any such principle must be valid in all epistemic states and since we have enormous freedom in choosing these three components, it is not surprising that only weak principles, such as $O(A|C) \Longrightarrow \neg O(\neg A|C)$, survive the test. Among the few that do survive, we find the sure-thing principle:

$$O(A|C) \wedge O(A|\neg C) \Longrightarrow O(A) \qquad (32)$$

read as "If you ought to do $A$ given $C$ and you ought to do $A$ given $\neg C$, then you ought to do $A$ without examining $C$". But one begins to wonder about the value of assembling a logic from a sparse collection of such impoverished survivors when, in practice, a full specification of $\kappa$, $\mu$, and $\Gamma$ would be required.

## 6.2 COUNTERFACTUAL CONDITIONALS

Stalnaker (1972) was the first to make the connection between actions and counterfactual statements, and he proposed using the probability of the counterfactual conditional (as opposed to the conditional probability, which is more appropriate for indicative conditionals) in the calculation of expected utilities. Stalnaker's theory does not provide an explicit connection between subjunctive conditionals and causation, however. Although the selection function used in the Stalnaker-Lewis nearest-world semantics can be thought of as a generalization of, and a surrogate for, causal knowledge, it is *too* general, as it is not constrained by the basic features of causal relationships such as asymmetry, transitivity, and complicity with temporal order. To the best of my knowledge, there has been no attempt to translate causal sentences into specifications of the Stalnaker-Lewis selection function.[9] Such specifications were partially provided in (Goldszmidt & Pearl 1992), through the imaging function $\omega^*(\omega)$, and are further refined in this paper by invoking the persistence model (Eq. (19)). Note that a directed acyclic graph is the only ingredient one needs to add to the traditional notion of *epistemic state* so as to specify a causality-based selection function.

From this vantage point, our calculus provides, in essence, a new account of subjunctive conditionals that is more reflective of those used in decision making. The account is based on giving the subjunctive the following causal interpretation: "Given $C$, if I were to perform $A$, then I believe $B$ would come about", written $A > B|C$, which in the language of ranking function reads

$$\kappa(\neg B|C) = 0 \text{ and } \kappa_A(\neg B|C) > 0 \qquad (33)$$

The equality states that $\neg B$ is considered a serious possibility prior to performing $A$, while the inequality renders $\neg B$ surprising after performing $A$. This account, which we call *Decision Making Conditionals* (DMC), avoids several paradoxes of conditional logics (see Nute 1992) and is further described in (Pearl 1993).

## 6.3 OTHER DECISION THEORETIC ACCOUNTS

Poole (1992) has proposed a quantitative decision-theoretic account of defaults, taking the utility of $A$, given evidence $e$, to be

$$\mu(A|e) = \Sigma_\omega \; \mu(\omega, A) P(\omega|e) \qquad (34)$$

This requires a specification of an action-dependent preference function for each $(\omega, A)$ pair. Our proposal (in line with (Stalnaker 1972)) attributes the dependence of $\mu$ on $A$ to beliefs about the possible consequences of $A$, thereby keeping the utility of each consequence constant. In this way, we see more clearly how the structure of causal theories should affect the choice of actions. For example, suppose $A$ and $e$ are incompatible ("If the light is on $(e)$, turn it off $(A)$"), taking (34) literally (without introducing temporal indices) would yield absurd results. Additionally, Poole's is a calculus of incremental improvements of utility, while ours is concerned with substantial improvements, as is typical of ought statements.

---

[9]Gibbard and Harper (Gibbard & Harper 1980) develop a quantitative theory of rational decisions that is based on Stalnaker's suggestion and explicitly attributes causal character to counterfactual conditionals. However, they assume that probabilities of counterfactuals are given in advance and do not specify either how such probabilities are encoded or how they relate to probabilities of ordinary propositions. Likewise, a criterion for accepting a counterfactual conditional, given other counterfactuals and other propositions, is not provided.

Boutilier (1993) has developed a modal logic account of conditional goals which embodies considerations similar to ours. It remains to be seen whether causal relationships such as those governing the interplay among actions and observations can easily be encoded into his formalism.

# 7   CONCLUSION

By pursuing the semantics of ought statements this paper develops an account of qualitative decision theory and a framework for qualitative planning under uncertainty. The two main features of this account are:
1. Order-of-magnitude specifications of probabilities and utilities are combined to produce qualitative expected utilities of actions and consequences, conditioned on observations (Eq. (7)).
2. A single causal network, combined with universal assumptions of persistence is sufficient for specifying the dynamics of beliefs under any sequence of actions and observations (Eq. (9)).

### Acknowledgements

This work benefitted from discussions with Craig Boutilier, Adnan Darwiche, Moisés Goldszmidt, and Sek-Wah Tan. The research was partially supported by Air Force grant #AFOSR 90 0136, NSF grant #IRI-9200918, Northrop Micro grant #92-123, and Rockwell Micro grant #92-122.